\documentclass{article}

\usepackage[preprint]{nips_2018}
\usepackage[utf8]{inputenc} 
\usepackage[T1]{fontenc}    
\usepackage[colorlinks=false,urlcolor=blue]{hyperref}       
\usepackage{url}            
\usepackage{booktabs}       
\usepackage{amsfonts}       
\usepackage{nicefrac}       
\usepackage{microtype}      
\usepackage{graphicx}	   
\usepackage[footnotesize,center]{subfigure}  
\usepackage{natbib}
\usepackage{authblk}
\newsavebox{\measurebox}

\newcommand{\figref}[1]{Figure~\ref{#1}}
\newcommand{\tableref}[1]{Table~\ref{#1}}

\title{Hierarchical Multi-task Deep Neural Network Architecture for End-to-End Driving}

\author{
 	Jos\'e Solomon  \ \ \ \  Fran\c cois Charette  \thanks{Both authors are members of the Ford DeepDSP Team. Please contact the authors via fcharett@ford.com.} \\ \ \\
  Greenfield Labs\\
  Ford Motor Company\\ 
  Palo Alto, CA 94305 \\
}

\begin{document}

\maketitle

\begin{abstract}
	A novel hierarchical Deep Neural Network (DNN) model is presented to address the task of end-to-end driving. The model consists of a master classifier network which determines the driving task required from an input stereo image and directs said image to one of a set of subservient network regression models that perform inference and output a steering command. These subservient networks are designed and trained for a specific driving task: straightaway, swerve maneuver, tight turn, gradual turn, and chicane. Using this modular network strategy allows for two primary advantages: an overall reduction in the amount of data required to train the complete system, and for model tailoring where more complex models can be used for more challenging tasks while simplified networks can handle more mundane tasks. It is this latter facet of the model that makes the approach attractive to a number of applications beyond the current vehicle steering strategy.
\end{abstract}

\section{Introduction}

End-to-end (E2E) deep learning driving models have been an active field of research in recent times. Although these models can reference their earliest instances to the late 1980's \cite{ALVINN}, it has only been in the past five years that they have fostered intense development and research in both academic circles as well as in industry. This is partially a result of advances in computational hardware that can now realistically be embedded in a vehicle platform, and also due in large part by advances in deep learning architecture itself whereby applications in image recognition \cite{VGG}, image segmentation \cite{SegNet}, and detector/classifiers \cite{Yolo, SSD} have been prime motivators in the effort to develop novel network components and graphs. Regarding the E2E task itself, the literature has a number of solid examples that have built on \cite{ALVINN}, beginning with the widely implemented model presented in \cite{NVIDIA} which uses a deep learning model to produce steering proposals based on image sensor data. Interesting work \cite{IntegratingProb} further builds on the E2E model given in \citep{NVIDIA} by embedding a modified version of its architecture into a probabilistic framework to optimize vehicle trajectory and provide an additional layer of quality verification for predicted steering output.

There are two fundamental facets to the challenge of creating robust E2E models: the model development task and the harvesting of training data. Although often the literature focuses on model architecture, (number of layers, layer types, etc.), it is not a decoupled system and often researchers are required to acquire large amounts of data to provide models with sufficient representation of the driving environment to impart the models with sufficient robustness. The model in \cite{NVIDIA} reports to have collected driving data in a number of different states and in a number of different environmental conditions, tallying a total of more than 70 hours of recorded driving data. This is a challenging task, especially when attempting to systematically represent a wide number of environmental conditions, (e.g. rain, snow, etc), and driving theaters, (interstate versus neighborhood driving), and to do so consistently across distinct test vehicles.

The conventional approach of developing E2E models is thus a combination of instilling a network architecture with enough layers, and so in turn sufficient weights (i.e. tunable parameters), to enable adequate vehicle response to any driving condition it encounters, as well as providing enough training examples to adequately refine said weights so that the network is an accurate inference engine. An implicit effect of such a one-model approach is that the most complex driving situations such as traffic intersections and hair-pin turns become the prime design criteria for network design and data harvesting, while the relatively straightforward tasks of lane keeping and straightaways become periphery elements of the training regime and are not core to the E2E model itself or corresponding training set. 

A key issue is this may create a Pareto frontier of competing objectives where the model strives to create optimized turning behavior at the expense of optimal lane-keeping behavior. What is proposed here is to make the E2E task modular in nature, and in so doing allow for the creation of dedicated regression models with more intricate network architecture along with sufficient training data for the more complex driving tasks, while using more simplified networks of fewer parameters and requiring less associated training data to manage the simpler aspects of the driving exercise. 

The current approach draws some inspiration from Multi-Task Learning (MTL) presented in the literature \cite{MTL}, where either a fixed set of layers form a shared representation of the input data and produce multiple task outputs, (i.e. hard sharing), or distinct parallel networks feed specific output tasks while sharing information among themselves, (i.e. soft sharing). MTL has been applied to E2E tasks in interesting ways, where \cite{Zipser} used an interesting multi-modal input in which stereo images along with a \textit{behavioral} modes are fed into a hard sharing MTL to produce steering and throttle control. More along the category of soft sharing, \cite{UDACITY} uses two distinct deep learning models, one for steering inference based on image input and one for throttle control using speed sequences as input, which share data across specific layers. The current work is somewhat correlated to the hard sharing paradigm, but uses a framework that is hierarchical in nature to allow for greater tailoring of network architecture as it is applied to a specific task. 

\section{The Hierarchical Multi-Task DNN Model}

\begin{figure}[htbp!]
  \centering
  \includegraphics[scale=0.4]{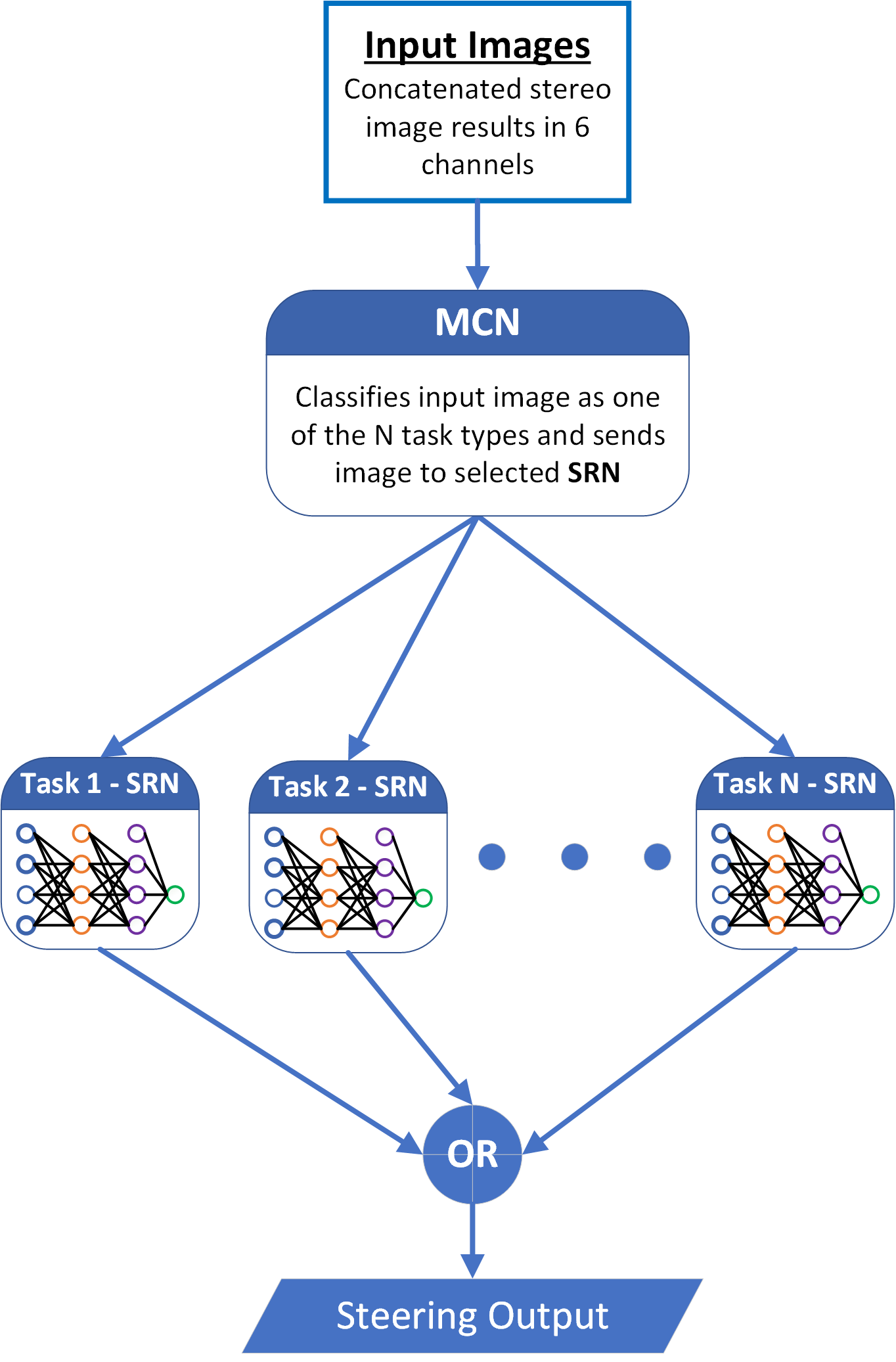}
  \caption{Hierarchical multi-task network model}
  \label{fig:two-level}
\end{figure}

The network architecture proposed is a two-level hierarchical system involving an over-arching classifier model, referred to as the Master Classifier Network (MCN), and a number of subservient regression models, referred to as Servant Regression Networks (SRN). The overall architecture is illustrated in \figref{fig:two-level}. The inference pathway is as follows: a stereo image is captured from the camera sensor and concatenated to create a six-channel entity. This entity is processed by the MCN, which classifies the image as a specific driving task. Based upon this classification, the image is passed on to a selected SRN, which then produces an inference based on it, producing an appropriate steering angle to feed to the vehicle's drive controller.

\newpage

Each SRN operates independently of its counterparts so that each image is processed only by the SRN selected by the MCN as the appropriate inference medium for the determined driving task. 

There are a number of performance characteristics specific to this modular approach. First, the system allows for tailoring the learned behavior of the model where distinct driving tasks are handled by distinct SRNs with their own training data sets. This is a key benefit since this in turn may lead to a significant overall reduction in the training data required for robust model performance. 

\begin{figure}[htbp!]
  \centering
  \includegraphics[scale=0.39]{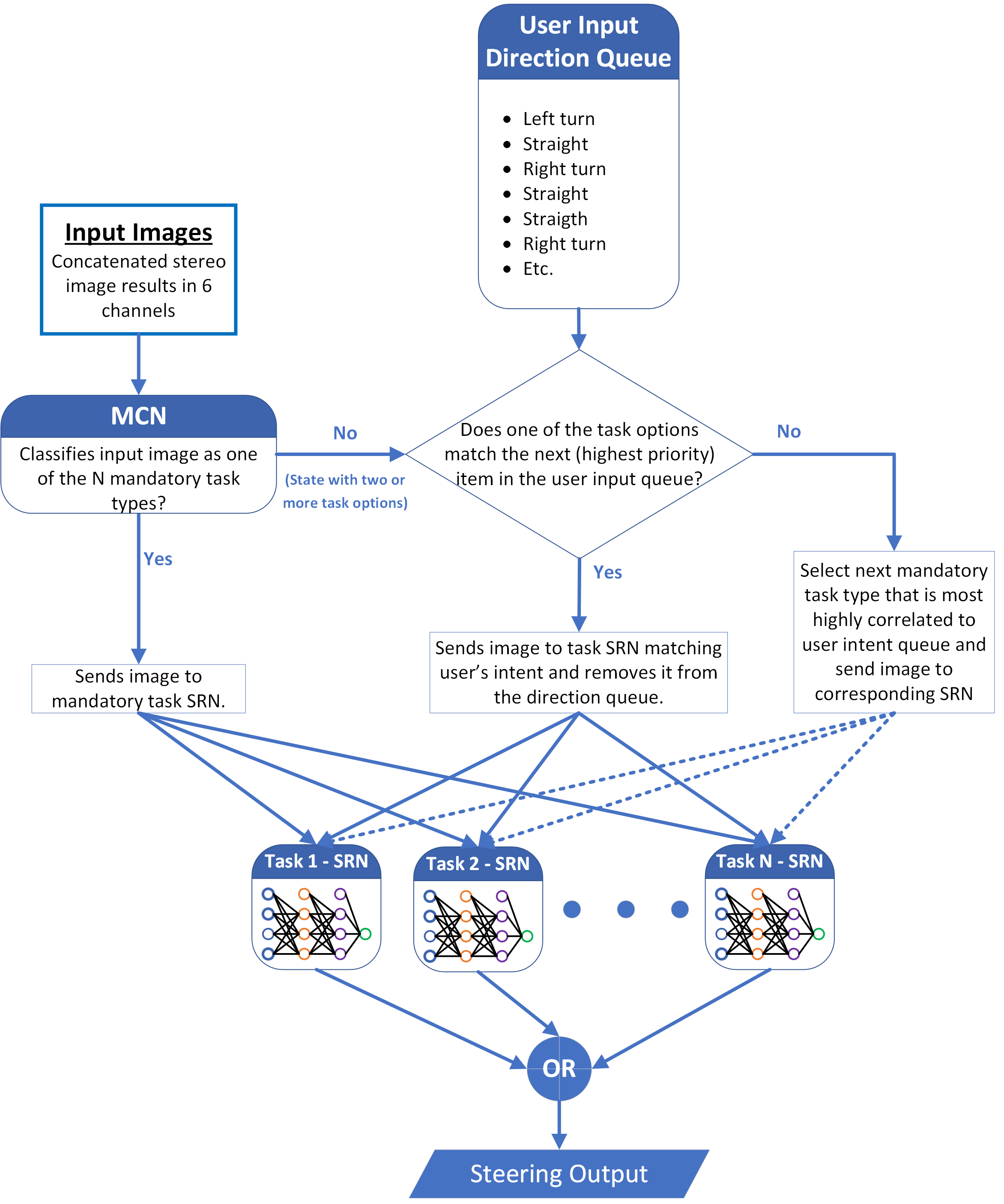}
  \caption{Decision engine with input from both the user intent queue and the MCN option set}
  \label{fig:multiTaskUserIntent}
\end{figure}

In addition, using this modular framework may allow future implementations of the system to place the intent of the user, (e.g. go from A to B in the least amount of traffic), to be placed at a more strategic juncture in the decision stream of the E2E model. In real-world driving applications, at different points in the drive path of the vehicle, not one but two or more driving tasks may be identified by the MCN as potentials for SRN processing. In those cases, a decision engine can be used to correlate user intent to the available driving task options, and in so doing select the task that is most highly coupled to the driver's objective. This framework is illustrated in \figref{fig:multiTaskUserIntent}.

\newpage

The work presented here is to explore the viability of this user-intent framework by implementing the system shown in \figref{fig:two-level} with the objective of (a) validating whether the MCN can be a reliable and efficient classification source for driving tasks, and (b) test if the overall hierarchical system can operate with the same degree of robustness as the single-model paradigm.

\subsection{Scaled-Vehicle Platform}

\begin{figure}[htp]
  \centering
  \subfigure{\label{xMaxx_1}\includegraphics[scale=0.039]{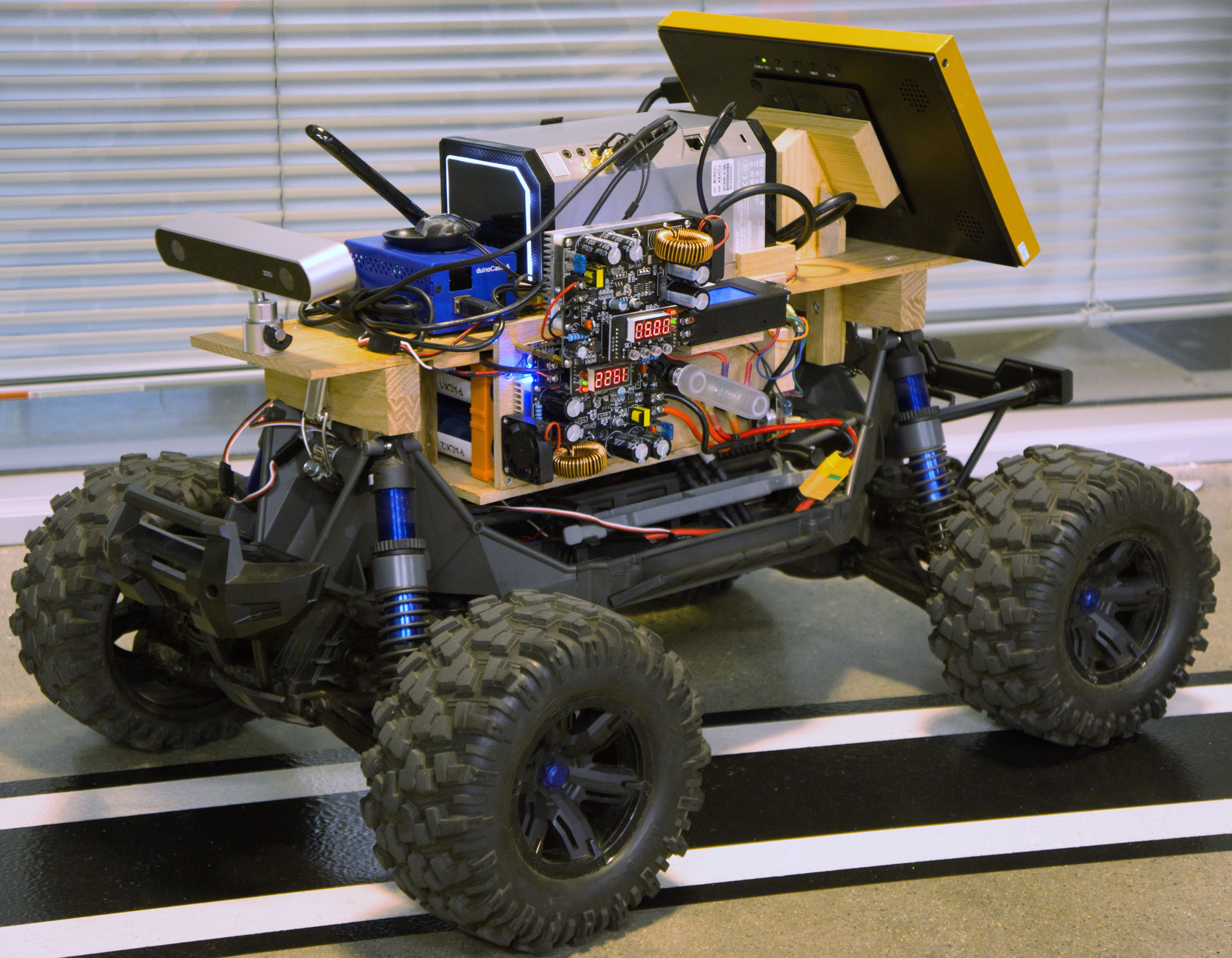}}
  \subfigure{\label{xMaxx_2}\includegraphics[scale=0.0405]{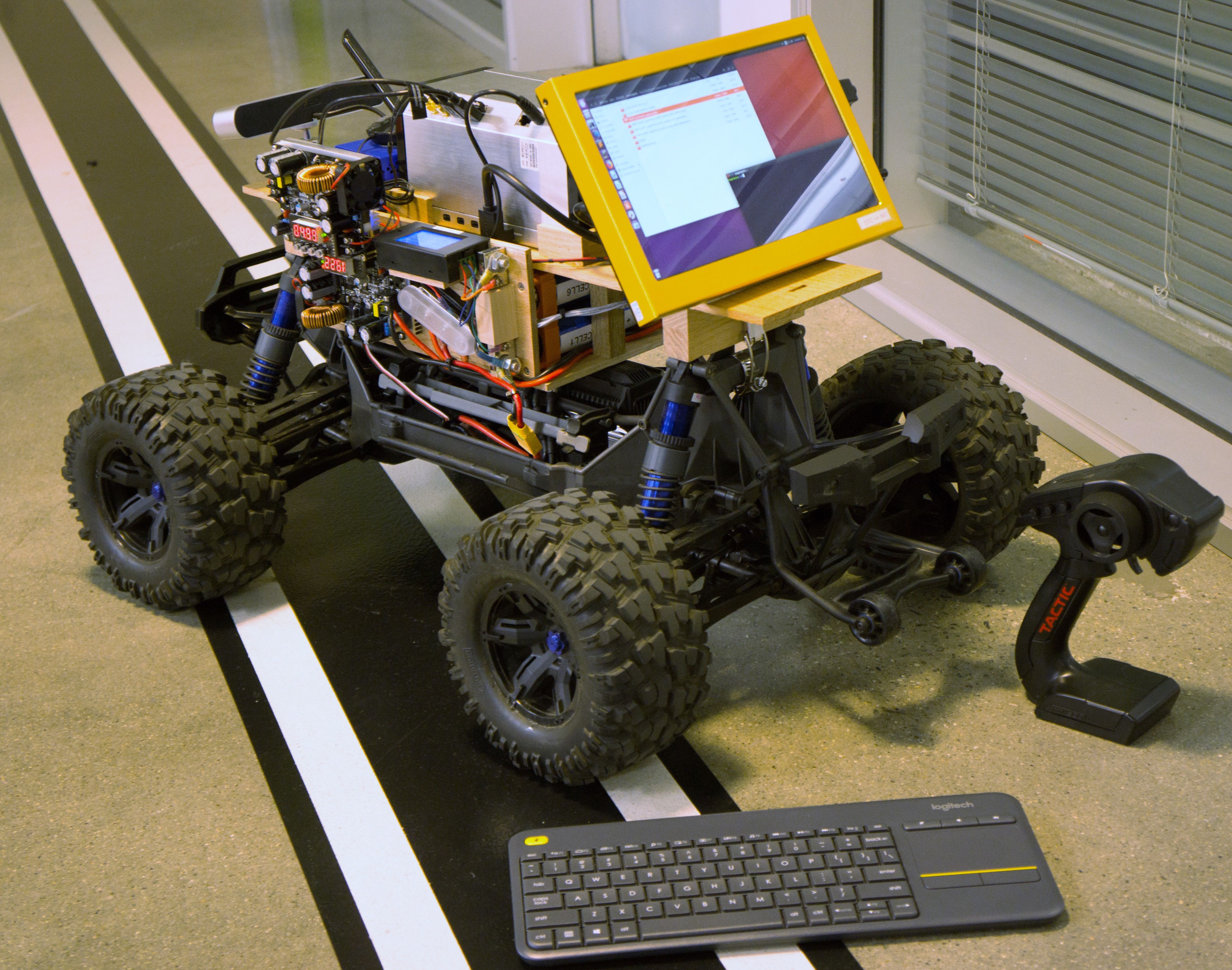}}
  \caption{Scaled-vehicle platform consisting of 1/6$^{th}$ scaled platform with Zed stereo-camera, Arduino Mega 2560 micro-controller, NUC running Ubuntu 16.04, keyboard and wireless transmitter to record steering input from driver}
  \label{fig:xMaxx}
\end{figure}

The scaled vehicle used for the work is based off a 1/6$^{th}$ scaled RC platform, where to the stock chassis, steering and suspension system, four distinct elements are added to permit the vehicle to operate autonomously: a stereo-vision Stereolab's Zed camera, an Arduino Mega 2560 Rev. 3 micro-controller, a NUC Intel i-7 PC with a discrete GTX-950 graphics card with 4 GB of memory, and a Vedder electronic speed controller (VESC) to manage the throttle level of the vehicle. The assembled vehicle is shown in \figref{fig:xMaxx}.

The vehicle has three driving modes: manual, recording and autonomous. In manual mode, the vehicle behaves as a standard RC car, with full throttle and steering control by the driver and no recording from the image sensor. In recording mode, the vehicle records the stereo image data, while logging the throttle speed, the operational frame rate of recording, and the steering angle input from the driver via the remote transmitter. This final entry is coupled with the image data from the stereo-camera to produce the training data for the SRNs. All data is stored on a frame-by-frame basis; i.e. the steering input is saved with its corresponding image set, throttle and frame rate as a single data entry within a HDF5 log file \cite{HDF5}.  

Using a post processing editor developed for the work, the raw Pulse Wave Modulation (PWM) steering input recorded during vehicle operation is normalized from -100 (a full left turn) to 100 (a full right turn) and the images themselves are resized from 672$\times$376 to 168$\times$94. In addition, so as to train the MCN, each frame is labeled as one of $N$ driving tasks. Aspects of this labeling will be discussed in the next section. Once the data is labeled, a copy is made which is then placed into separate training sets, where each set is dedicated to the training and testing of specific SRNs. In this way, the training data is used twice: first as a labeled data set for the MCN, then as segregated regression data using the normalized steering angle for the SRNs. It is noted that in this supervised training framework, the sole input is the concatenated image frame, and the ground truth is either a driving task label or a normalized steering angle.

\subsection{Data Labeling and the Five Zone Track}

\begin{figure}[htp]
  \centering
  \includegraphics[scale=0.1]{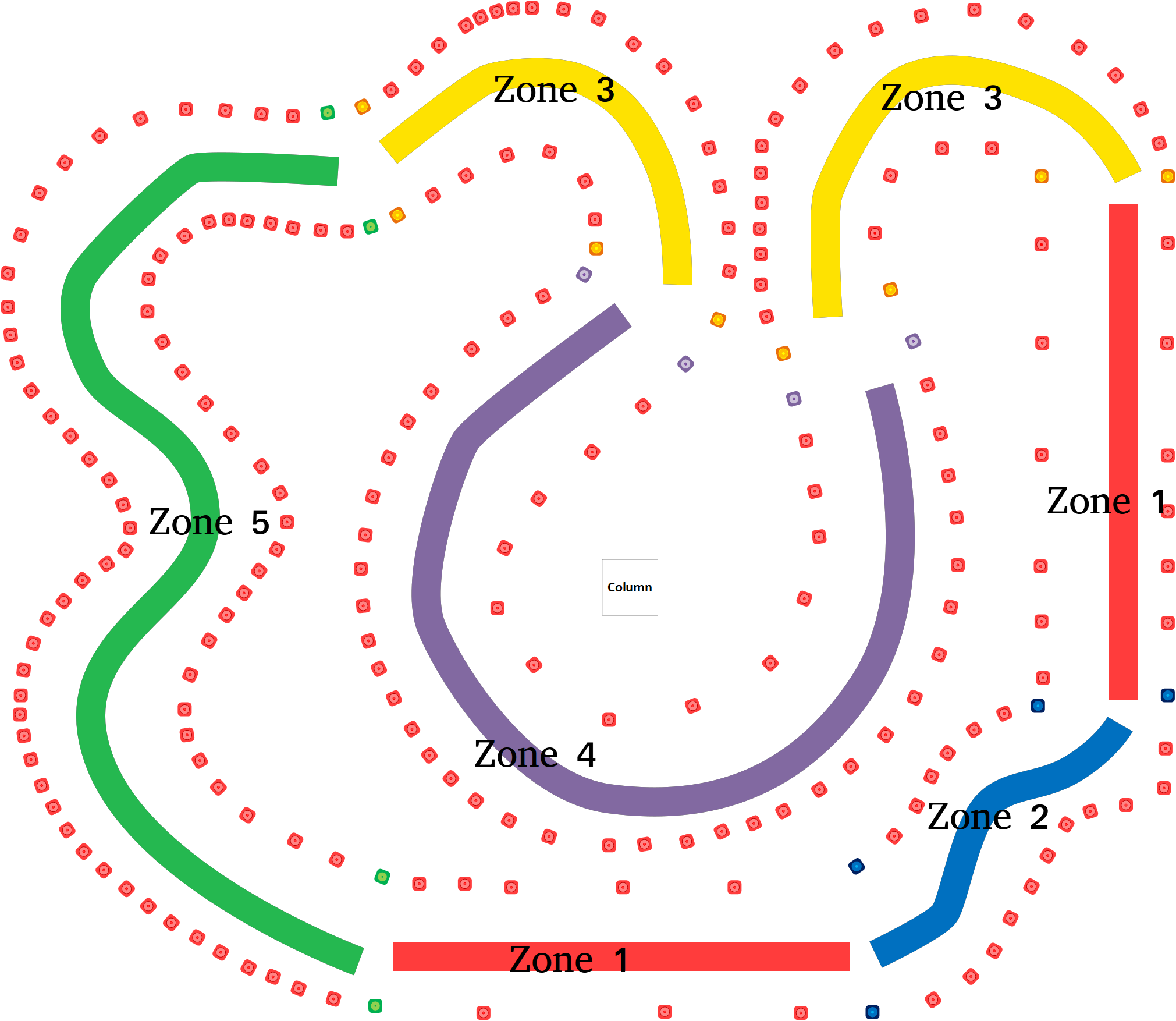}
  \caption{Five zone track}
  \label{fig:5zone}
\end{figure}

A specialized track was created for the current work. Using 7-inch cones of two differing shades of orange, the track illustrated in \figref{fig:5zone} was established as the test bed. As noted in the figure, there are five distinct zones that are representative of different driving tasks.

\begin{itemize}
\item[] Zone 1: Straightaway sections of the track
\item[] Zone 2: Swerve maneuver connecting Zone 1 sections
\item[] Zone 3: Tight turn occurring at the entry and exit of Zone 4
\item[] Zone 4: Gradual turn occurring at the center of the track
\item[] Zone 5: Chicane portion
\end{itemize}

When harvesting data for training, the scaled vehicle is driven in both clock-wise and counter clock-wise direction an equal number of laps so as to permit the vehicle to negotiate the track regardless of its direction of travel. Using the post processing editor previously mentioned, a zone label is added to the data that is logged by the vehicle, which is subsequently used to train the MCN.

\section{Network Models}

This section presents the deep learning models used for the MCN and for the five distinct SRNs. Each network was designed to be as shallow as functionally possible so as to reduce the memory footprint of weights as well as the active memory required during inference. All models were designed, trained and deployed using the Tensorflow \cite{Tensorflow} framework.

\subsection{Loss Functions}

Two distinct cost functions were used: softmax with cross entropy loss for the MCN, with the square of the L2 loss function for the SRNs. 

The MCN consists of a five-zone classifier, where each driving zone is to be classified as a distinct driving task. The  formulation for the softmax categorical function is given as a probability vector where each component of the vector is given as

\begin{equation}
p_i \; = \; \frac{e^{v_i}}{\sum\limits_{k=1}^M e^{v_k}}
\end{equation}

The corresponding cross-entropy loss is given as 

\begin{equation}
H(y, p) \; = \; - \sum\limits_j y_j \; \log(p_j)
\end{equation}

where $y_j$ is the ground truth label of the given training sample $j$.

Each SRN is a regression model built on the normalized steering recorded during logging of training data. As such, the loss function selected was Tensorflow's \textit{l2\_loss}, which as per Tensorflow documentation \cite{Tensorflow} is defined as half the square of the $L_2$ norm, giving

\begin{equation}
\frac{1}{2}||(\mathbf{y} - \mathbf{\hat{y}})||_2^2 \; = \; \frac{1}{2} \sum\limits_{k=1}^N (y_k \; - \; \hat{y}_k)^2
\end{equation}

where $\mathbf{\hat{y}}$ is the steering model output of a given SRN. 

\subsection{Layers}

In terms of the network architecture of the MCN and SRNs, for the most part these are based on alternating implementations of three fundamental deep learning layer types: convolutional layers, fully-connected layers and Recurrent Neural Network (RNN) layers. Key parameters for each applied layer are given in the network layout illustrations which follow. The convolutional layers are based on the cross-correlation operator

\begin{equation}
(f * g) [n] \; = \; \sum\limits_{m=-\infty}^{\infty}f[m]g[m+n]
\end{equation}

where the applied kernel dimensions, application stride and depth were varied from network to network. Fully-connected layers,

\begin{equation}
H(x) \; = \; W^T \cdot x \; + \; b
\end{equation}

are used on a one-layer per model basis to minimize memory footprint, while RNNs,

\begin{equation}
H(x) \; = \; (W^T \cdot x + W_{h}^T \cdot \sigma(h_{t-1})) \; + \; b
\end{equation}

where applied as unidirectional operators, (using only \textit{past} instances), and used in conjunction with fully-connected layers only in one SRN model.

ReLU is the activation function used throughout all models.

\subsection{Architecture}

\begin{figure}[htbp!]
  \centering
  \includegraphics[scale=0.4]{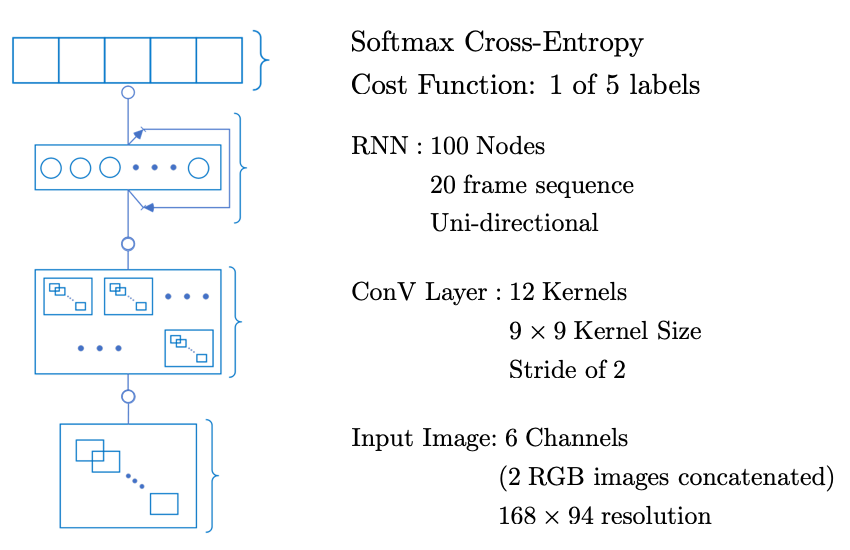}
  \caption{MCN Model}
  \label{fig:MCN}
\end{figure}

The MCN is a primary component of the mult-task framework, and through various model iterations, a robust model resulted from the straightforward architecture given in \figref{fig:MCN}.

It is noted that the concatenated image that is the input to the network is passed as-is and is unaltered when sent to the subsequent SRN selected by the MCN as the appropriate network for the determined driving task. 

For each of the SRN layouts that follow, the design emphasis is to create as shallow a network as possible that is able to robustly perform its respective driving task. As such, each of the models is a result of iteratively training networks then testing them on test data frames not used for training, and if performing reliably on those, actual driving tests on the track. A model is labeled \textit{successful} if it negotiates its respective section of track three times in both clockwise and counter-clockwise direction without moving a cone. As such, each model that is presented next contributed to the multi-task framework which allowed the vehicle to complete three laps of the track in both directions fully autonomously.

\begin{figure}[htbp!]
  \centering
  \includegraphics[scale=0.4]{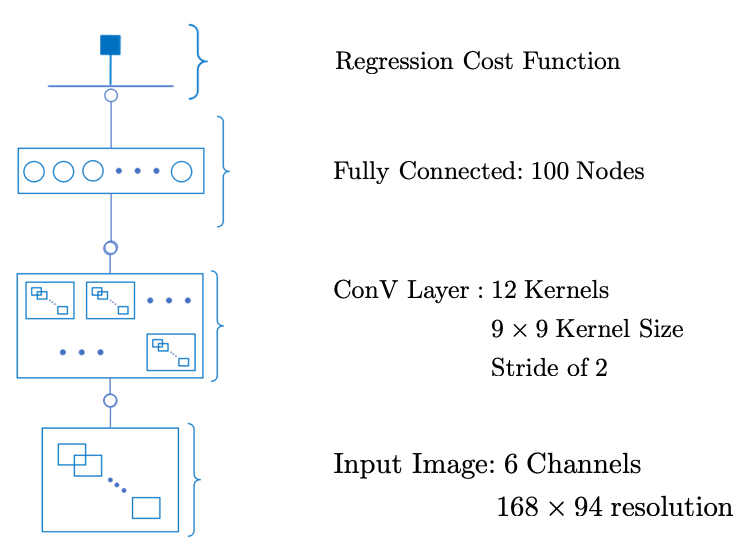}
  \caption{SRN Zone 1 Model - Straightaway sections}
  \label{fig:Zone_1}
\end{figure}

The model for Zone 1, (the straightaway sections of track), is presented in \figref{fig:Zone_1}.

Given that the driving task is to maintain the vehicle straight within a lane of cones, a simple network was readily designed.

\begin{figure}[htbp!]
  \centering
  \includegraphics[scale=0.4]{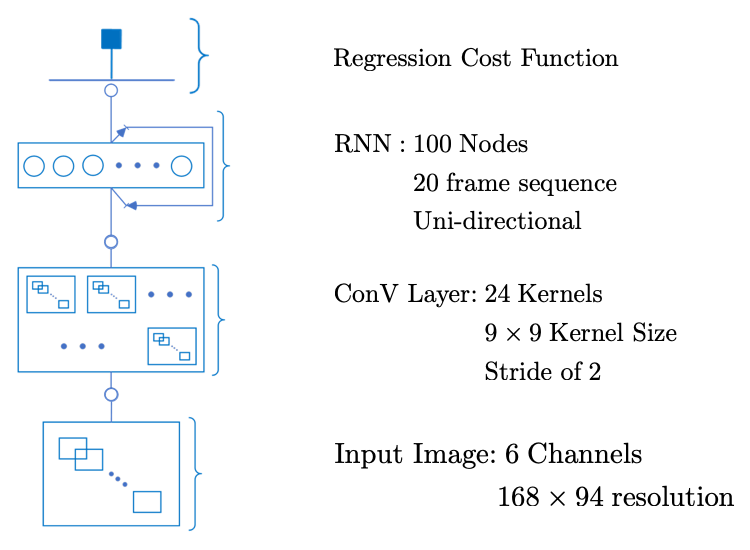}
  \caption{SRN Zone 2 Model - Swerve Maneuver}
  \label{fig:Zone_2}
\end{figure}

Zone 2 is a much more challenging task since the vehicle is forced to span its full range of steering in a relatively short travel distance, thus mimicking a swerve maneuver. Initial attempts of doing so with additional convolutional layers or a series of fully-connected layers towards the end of the network produced bloated models with limited reliability. State memory was incorporated using an RNN layer, with the fine tuning of 20 frames in the \textit{past} tense proving optimal. The resulting network is shown in \figref{fig:Zone_2}.

\begin{figure}[htbp!]
  \centering
  \includegraphics[scale=0.4]{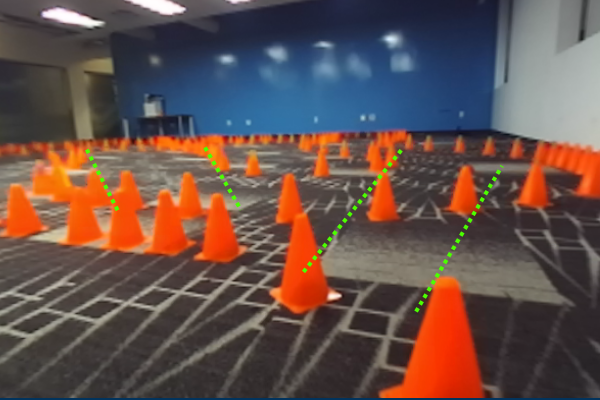}
  \caption{Zone 3 scaled-vehicle perspective highlighting false route alternatives}
  \label{fig:Zone_3_confusion}
\end{figure}

Zone 3 is dynamically challenging given the tight steering maneuver required and at a certain point in the turn, cones from other sections of the track enter the visual scope of the vehicle and produce confusing false-alternatives. (In certain instances and with certain network architectures, the vehicle drifted into those trajectories during trial runs.) A frame grab from the perspective of the vehicle is shown in \figref{fig:Zone_3_confusion}.

\begin{figure}[htpb!]
  \centering
  \includegraphics[scale=0.4]{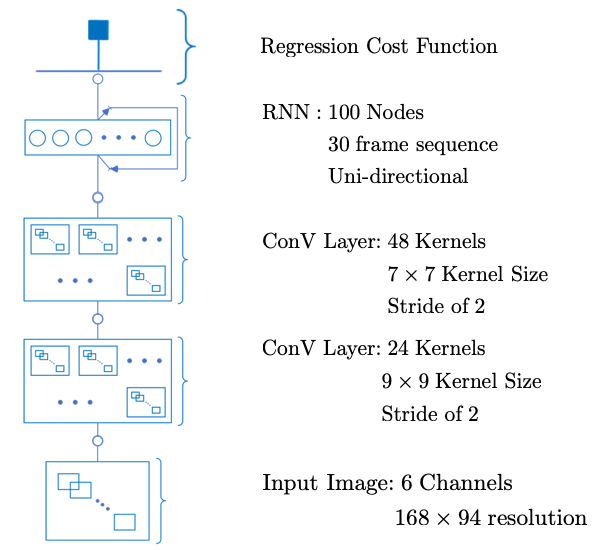}
  \caption{SRN Zone 3 Model - Tight Turn}
  \label{fig:Zone_3}
\end{figure}

As a result, a second convolutional layer is added to the Zone 2 SRN, creating greater depth and thus more feature maps to create greater points of reference for the correct trajectory. Furthermore, an RNN layer with slightly more frame history, (30 frame instances vs. 20 for the Zone 2 SRN), is connected to the output of the second convolutional layer. The final model is illustrated in \figref{fig:Zone_3}.

\newpage

\begin{figure}[htpb!]
  \centering
  \includegraphics[scale=0.4]{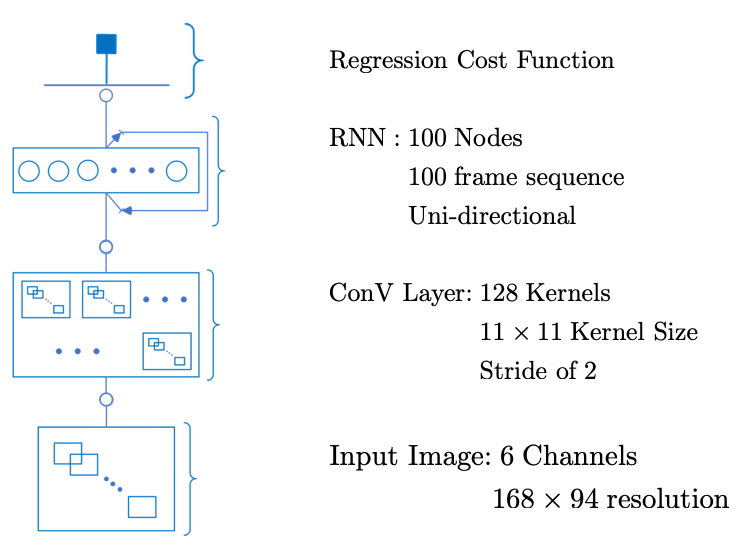}
  \caption{SRN Zone 4 Model - Gradual Turn}
  \label{fig:Zone_4}
\end{figure}

For Zone 4, the driving task is a straightforward gradual turn and so the key here is to minimize the confusion produced from cones from unrelated sections of the track. The Zone 4 SRN builds on the Zone 2 SRN by simply including a very deep initial convolutional layer and then using an RNN of a relatively long sequence length to ensure the vehicle trajectory builds on the correct features. The model is illustrated in \figref{fig:Zone_4}.

\begin{figure}[htpb!]
  \centering
  \includegraphics[scale=0.55]{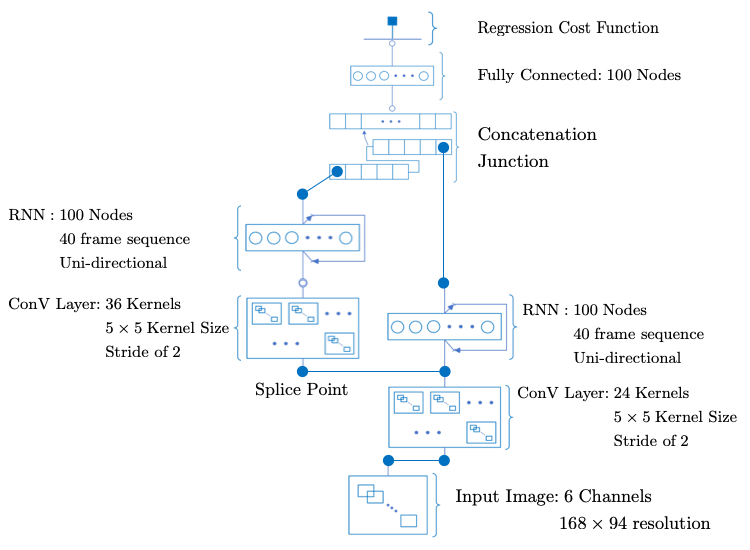}
  \caption{SRN Zone 5 Model - Chicane}
  \label{fig:Zone_5}
\end{figure}

\newpage
Zone 5 is the chicane, which is the most challenging driving task of the track and as such required more iterations of network design. The network's final form, presented in \figref{fig:Zone_5}, is a byproduct of studying the vehicle's visual perspective as it traveled through this section of track. 

\newpage

\begin{figure} [htpb!]
\centering
\begin{tabular}{cc}
\includegraphics[width=0.5\textwidth]{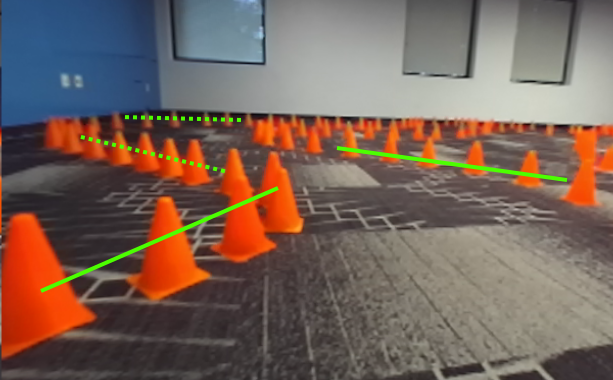} &
\begin{minipage}[b]{0.425\textwidth}
\includegraphics[width=1.0\textwidth]{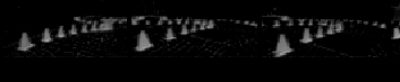} \\
{(b) First convolutional layer output (24 kernels shown in \figref{fig:Zone_5})} \\
\ \\
\includegraphics[width=1.0\textwidth]{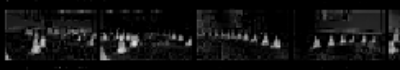} \\
 {(c) Second convolutional layer output (36 kernels shown in \figref{fig:Zone_5})}
\end{minipage} \\
{(a) Primary and secondary cones} &  \\[4pt]
\end{tabular}
\caption{SRN Zone 5 model vehicle visual scope and convolutional output samples, (solid lines are cone rows directly in the course of the vehicle, while dotted lines indicate secondary cone rows that are targeted features for the feature maps created by the second convolutional layer). (b) and (c) show samples of the feature maps created by the first and second convolutional layers respectively.}
\label{fig:Zone_5_dev}
\end{figure}

As noted in the network layout, the first convolutional layer splits its output to subsequent layers: one is a RNN with 40 sequence frames in the past sense; the other is a subsequent convolutional layer with increased depth, which in turn feeds to its own RNN layer, also of uni-directional past sequence recurrence. The two RNN layers then are concatenated and feed a fully-connected layer comprised of 100 nodes, which then feeds the output node. This \textit{split}-output design at the first convolutional layer came from noting that as the vehicle traverses the chicane, there is useful information in the cone row directly in front of the vehicle as well as in rows further downstream of the visual scope, as illustrated in \figref{fig:Zone_5_dev}.

Conceptually, for this driving task there is important information at various scales of the visual scope. Labeling primary cones as cones that are directly in the path of the vehicle, and secondary cones as cones that are within the driving zone, but not directly in the vehicle's path, (i.e. the vehicle needs to complete a turn before the cone is in its direct path), useful information can be extracted from both categories. The intent of the network design is for the first convolutional layer to capture most of the visual cues stemming from primary-to-primary cone interaction, and attaching state memory to these feature maps will assist the vehicle in navigating its immediate environment. The second convolutional layer, which has a larger receptive field since it is applied to the output of the first convolutional layer, aims to extract further information from the scene by monitoring primary-to-secondary cone interaction, which might only be evident in a larger spatial scale of the input image. An example is to note how quickly secondary cones move from left-to-right or right-to-left in relation to primary cones. This relative movement is then embedded in the second RNN layer, and a fully connected layer couples both memory states together. It is noted that this network proved adept at navigating the chicane and has created an architectural framework for separate work conducted outside of the controlled environment of the five zone track.

\subsection{Training}

Training the multi-task model begins with driving the scaled-vehicle around the track in recording mode. The two images captured from the stereo-camera are logged sequentially along with steering input from the driver. Each lap completed is stored as a separate log file, thus facilitating the editing and labeling tasks to follow.  

Training the MCN required the use of all the logged training data along with zone categorical labels. Each subservient SRN was trained on data frames extracted from this base lap set, where the extracted frames only document the specified driving task for which the SRN is responsible. As the SRNs were trained, the number of base laps, (the complete laps which frames were extracted from), was reduced from a total of 10 in each direction to 4 in cases were robust network performance was readily evident by monitoring the loss curve. This facet is documented in the \textit{Base Log Files} entry in \tableref{table:training_schedule}. Training epochs ranged from 100 to 300 once again based on the behavior of the loss curve. 

\begin{table}[htbp!]
  \caption{Training schedule for the MCN and 5 SRNs}
  \label{table:training_schedule}
  \centering
  \begin{tabular}{lllll}
    \toprule
    Model & Base Log Files & Training Data Frames & Test Data Frames & Epochs \\
    \midrule
	MCN        & 20 (10 cw / 10 ccw)*&  71,623                 & 28,572              & 100 \\
	SRN Zone 1 & 8 (4 cw / 4 ccw)    &   6,408                 &  6,237              & 200 \\
	SRN Zone 2 & 20 (10 cw / 10 ccw) &   6,235                 &  2,542              & 200 \\
	SRN Zone 3 & 20 (10 cw / 10 ccw) &  15,058                 &  6,102              & 200 \\
	SRN Zone 4 & 8 (4 cw / 4 ccw)    &   7,081                 &  7,184              & 300 \\
	SRN Zone 5 & 8 (4 cw / 4 ccw)    &   6,580                 &  6,507              & 200 \\
    \bottomrule \\
    \multicolumn{2}{c}{* cw: clockwise, ccw: counter-clockwise} 
  \end{tabular} \\
\end{table}

As noted in the table, the SRNs for Zone 1, 4 and 5 required only 8 base files, and as such, a higher number of test data frames were used to verify model performance before actual driving experiments.

\section{Results}

Two primary sets of results are documented. Separate test laps were logged during data acquisition, and these log files are used to test model performance before actual driving experiments. A total of eight logged laps are used for testing purposes, four in each direction. After the MCN and SRNs were tested independently, the multi-task framework as a whole is tested, producing a steering curve for entire laps logged by the test files.

It is noted that once a successful model is predicted by the test data set, it is important to conduct an actual driving experiment. This is critical given that when the vehicle is actual navigating the course, it may find itself deviating from the trajectory space of the training data and thus the actual recovery capability of the model, (i.e. returning to the center of the track), is not apparent or quantifiable by simply running on test data sets, which although separate from training data, are not dissimilar in terms of vehicle trajectory recorded. Thus a key metric to evaluate an end-to-end model is how capable the vehicle is in returning to an appropriate track trajectory if it deviates from training data at any point in the completion of a lap. A successful model is one that is able to complete three consecutive laps in both directions without moving a cone.

\subsection{Test Files Analysis}

\begin{figure}[htp]
  \centering
  \subfigure[Clockwise drive direction]{\label{cf_matrix_cw}\includegraphics[scale=0.325]{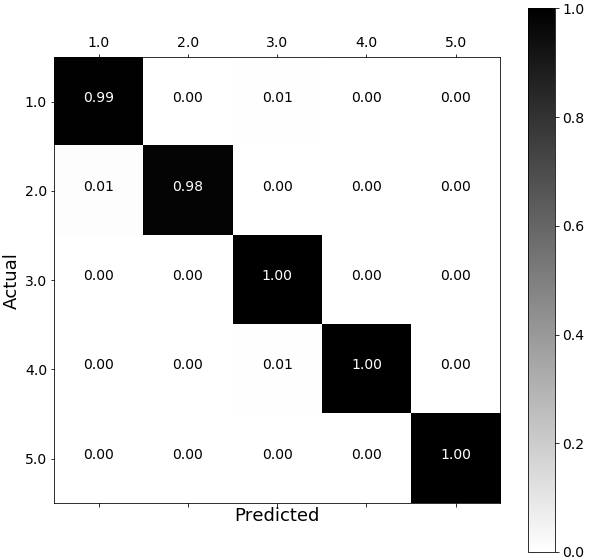}}
  \subfigure[Counter-clockwise drive direction]{\label{cf_matrix_ccw}\includegraphics[scale=0.325]{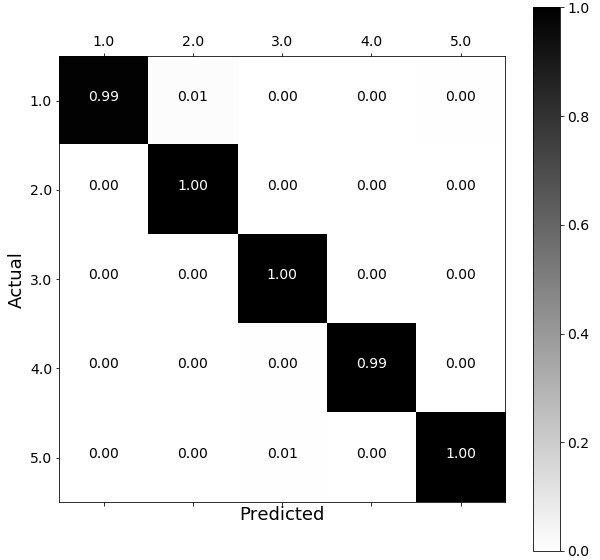}}
  \caption{Confusion matrix for MCN based on test log files.}
  \label{fig:confusion_matrix}
\end{figure}

The first element that is presented is the classification capabilities of the MCN. Traversing the track in both clockwise and counter-clockwise fashion using the same model is critical, and the MCN is capable of correctly classifying an image belonging to a given zone regardless of the direction of travel. Confusion matrices for the MCN based on test laps in both directions are shown in \figref{fig:confusion_matrix}.

\begin{figure} [htpb!]
\centering
\begin{tabular}{cc}
\includegraphics[width=0.5\textwidth]{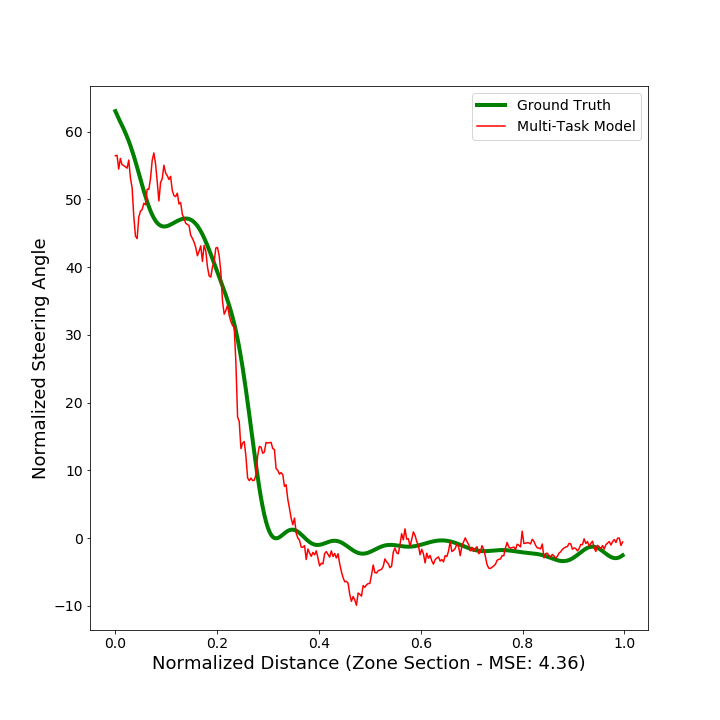} &
\includegraphics[width=0.5\textwidth]{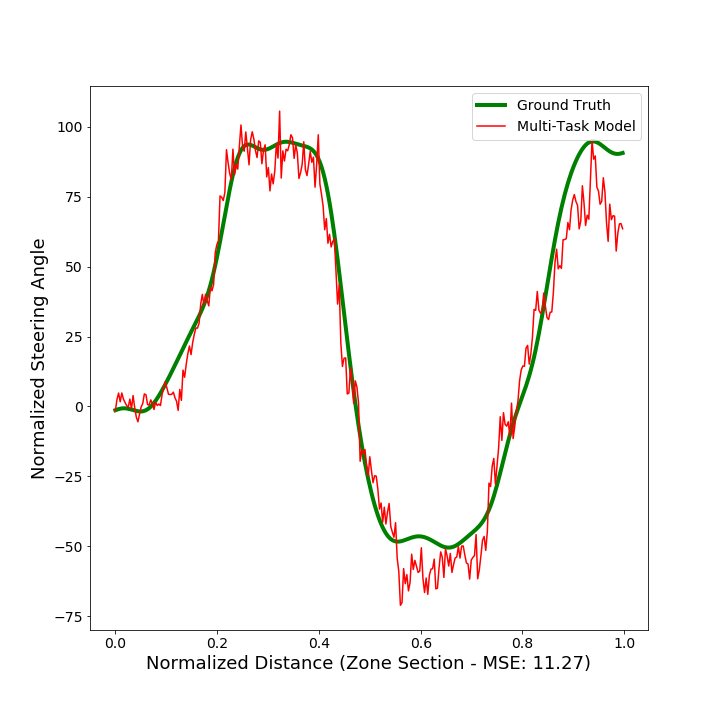} \\
{(a) Zone 1 SRN}  & {(b) Zone 2 SRN}  \\[4pt]
\end{tabular}
\begin{tabular}{cc}
\includegraphics[width=0.5\textwidth]{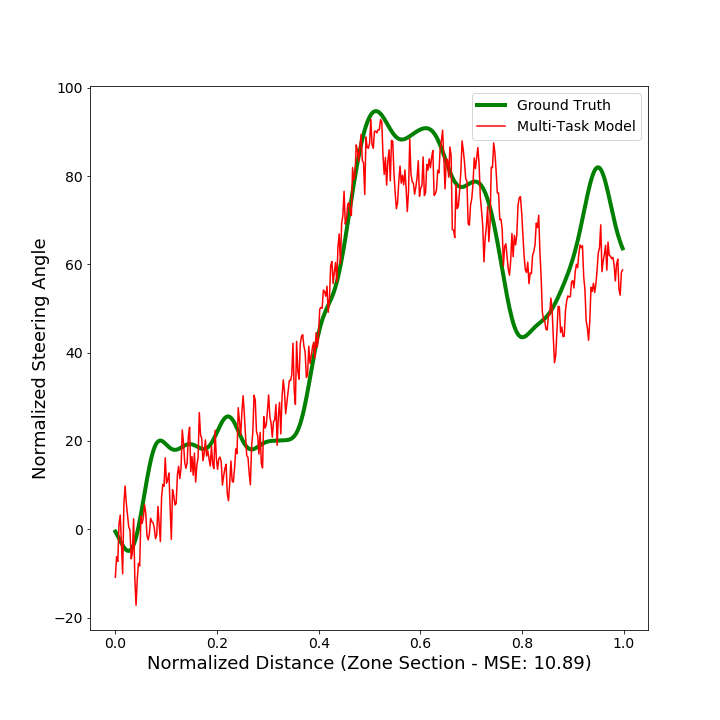} &
\includegraphics[width=0.5\textwidth]{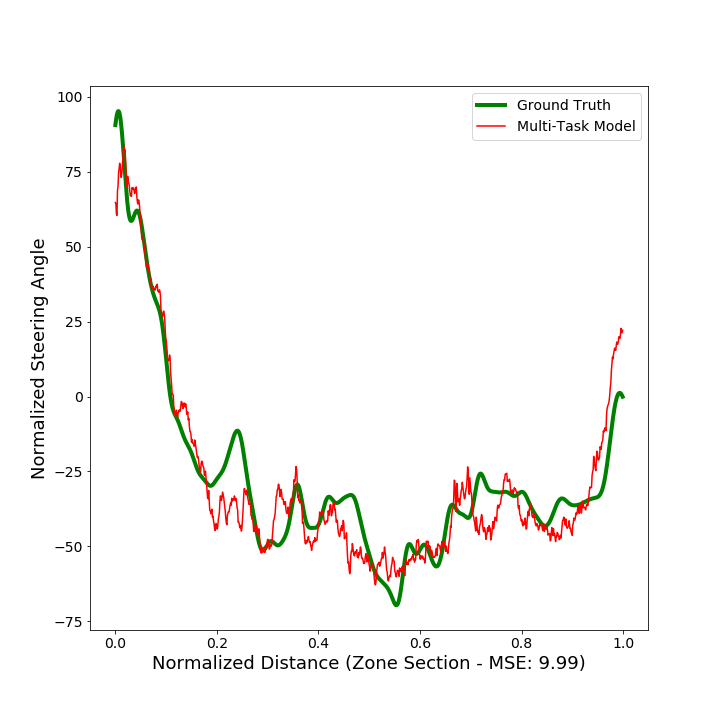} \\
{(c) Zone 3 SRN}  & {(d) Zone 4 SRN}  \\[4pt]
\end{tabular}
\begin{tabular}{cc}
\includegraphics[width=0.5\textwidth]{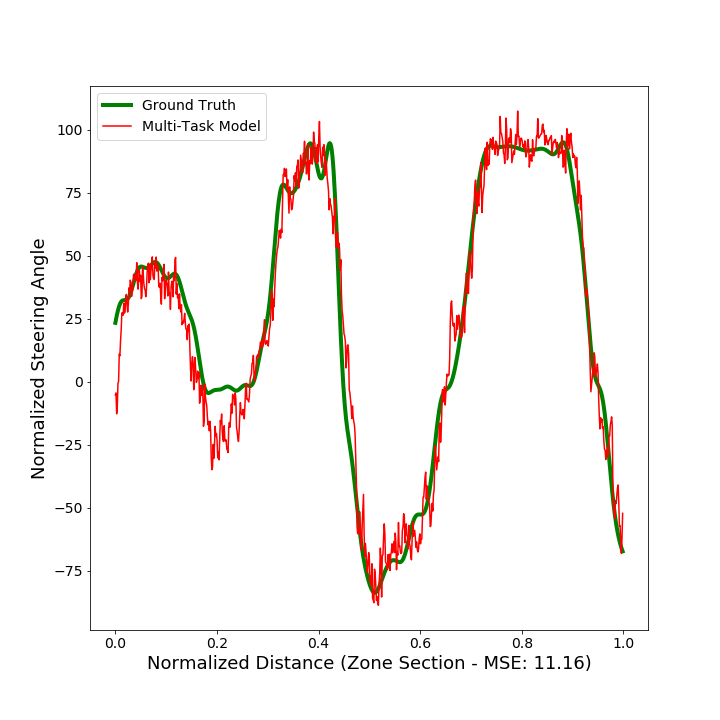} \\
{(d) Zone 5 SRN} \\[4pt]
\end{tabular}
\caption{SRNs sample steering predictions for clockwise lap direction.}
\label{fig:srn_cw}
\end{figure}

\begin{figure} [htpb!]
\centering
\begin{tabular}{cc}
\includegraphics[width=0.5\textwidth]{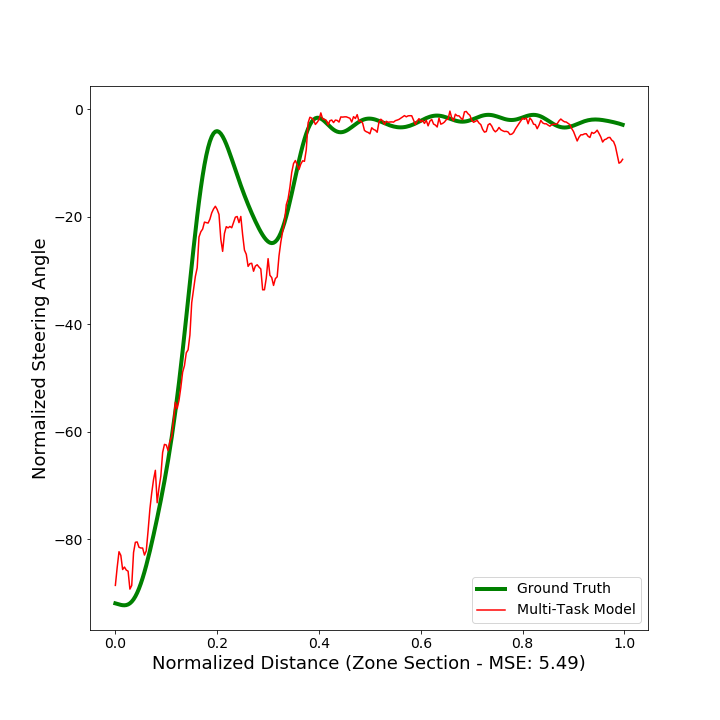} &
\includegraphics[width=0.5\textwidth]{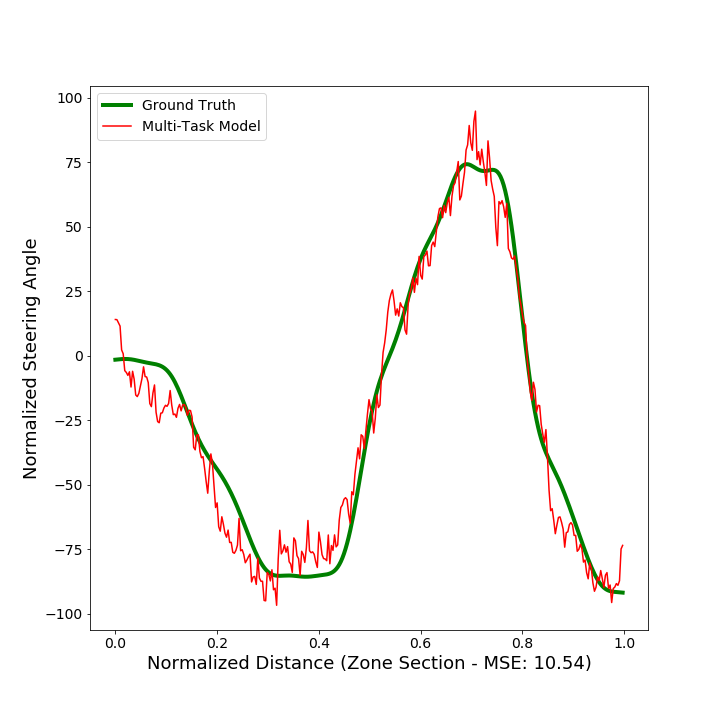} \\
{(a) Zone 1 SRN}  & {(b) Zone 2 SRN}  \\[4pt]
\end{tabular}
\begin{tabular}{cc}
\includegraphics[width=0.5\textwidth]{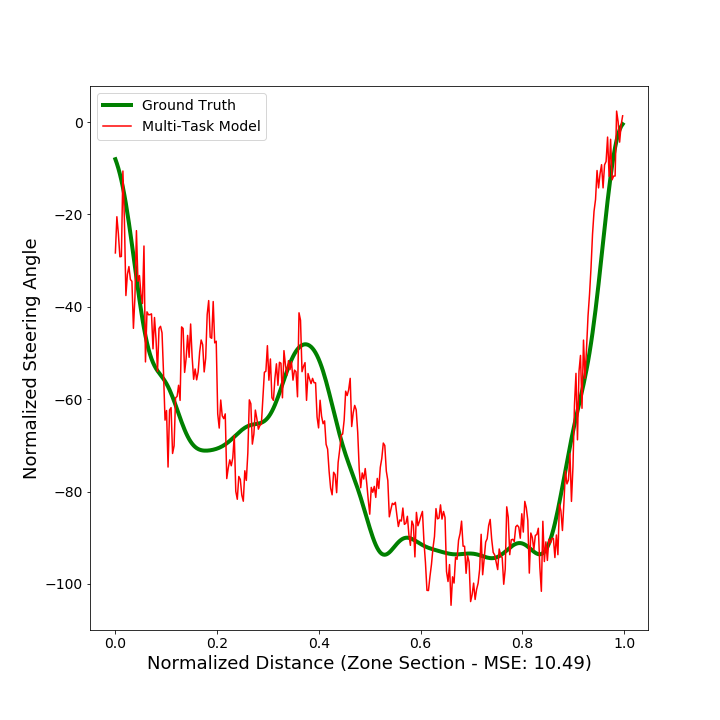} &
\includegraphics[width=0.5\textwidth]{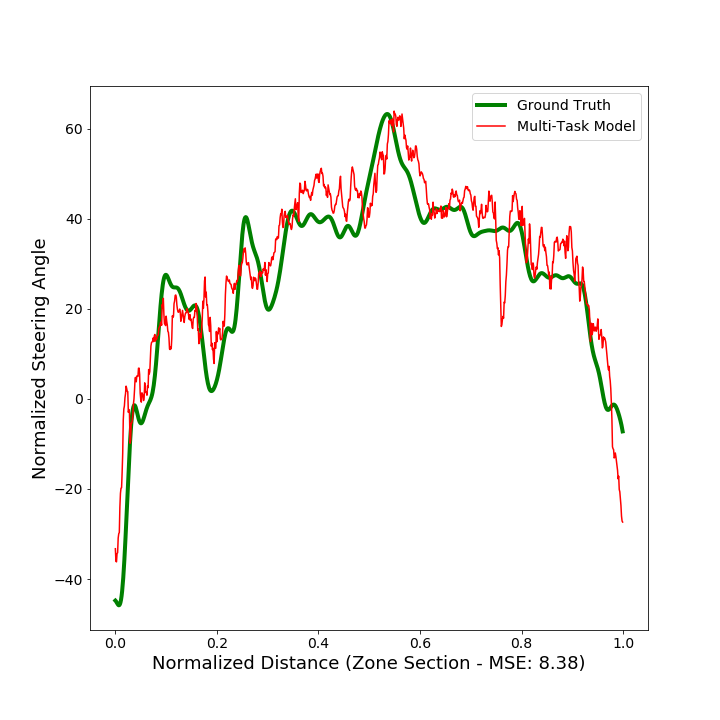} \\
{(c) Zone 3 SRN}  & {(d) Zone 4 SRN}  \\[4pt]
\end{tabular}
\begin{tabular}{cc}
\includegraphics[width=0.5\textwidth]{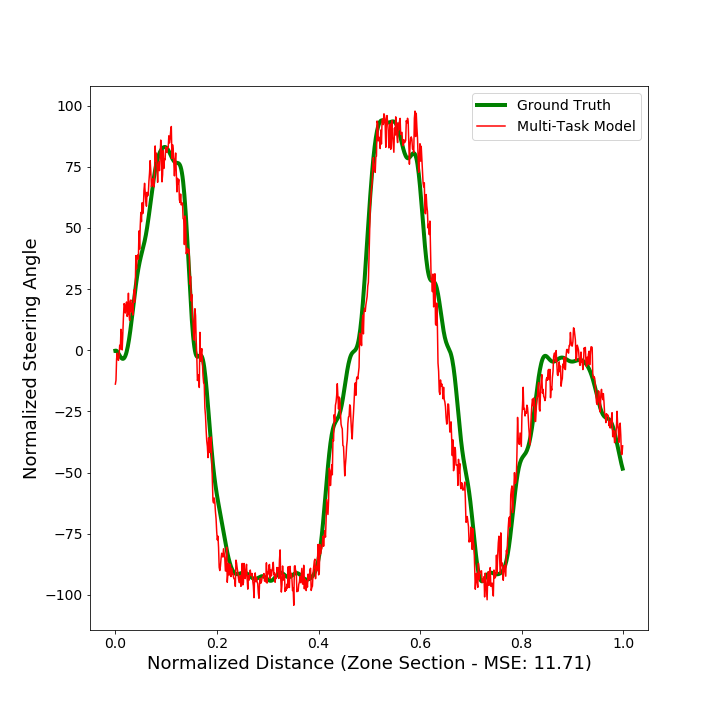} \\
{(d) Zone 5 SRN} \\[4pt]
\end{tabular}
\caption{SRNs sample steering predictions for counter-clockwise lap direction.}
\label{fig:srn_ccw}
\end{figure}

Moving from classification to regression, each SRN was tested on a data sets pertaining to each drive zone. A subset of the test data results is shown in \figref{fig:srn_cw} and \figref{fig:srn_ccw}, with sequential data frames used to illustrate each network's predicted steering response as compared to that of the ground truth, (which is the human driving trajectory).

As noted from the plots, Mean Squred Error (MSE) values range from roughly 4 to 12 depending on drive zone. Although an MSE value in itself is not an absolute determinant of model success, an MSE greater than 20, which translates to 10\% of the total steering range, would indicate further refinement of the model is required.

\begin{figure}[htp]
  \centering
  \subfigure[Clockwise drive direction]{\label{cw_test_log}\includegraphics[scale=0.325]{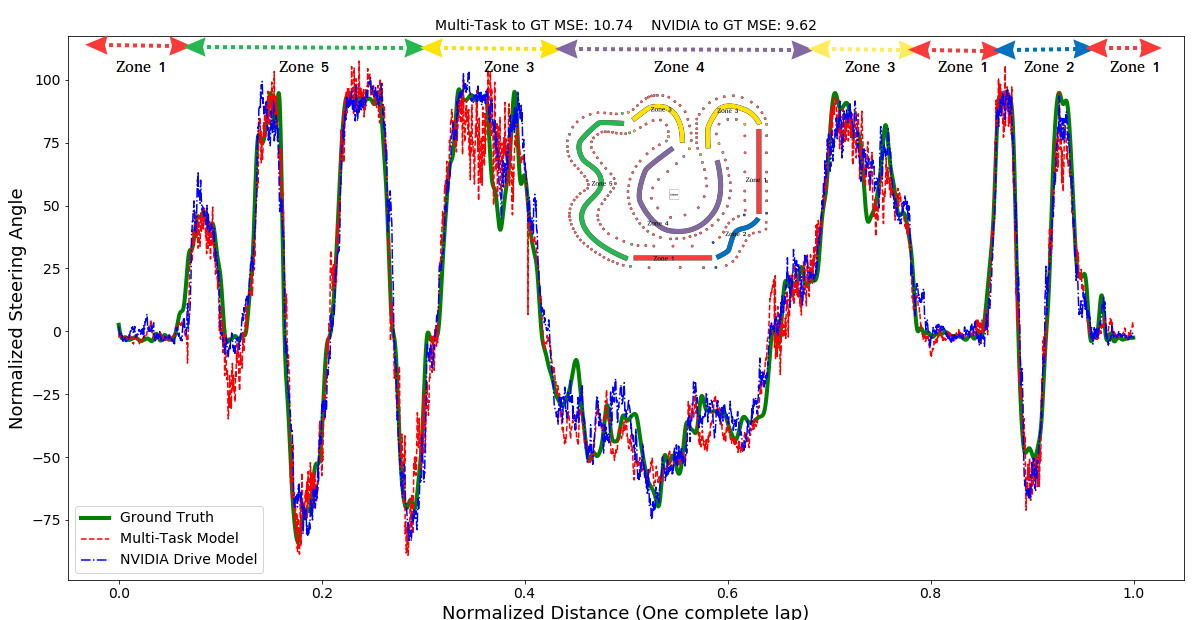}}
  \subfigure[Counter-clockwise drive direction]{\label{ccw_test_log}\includegraphics[scale=0.325]{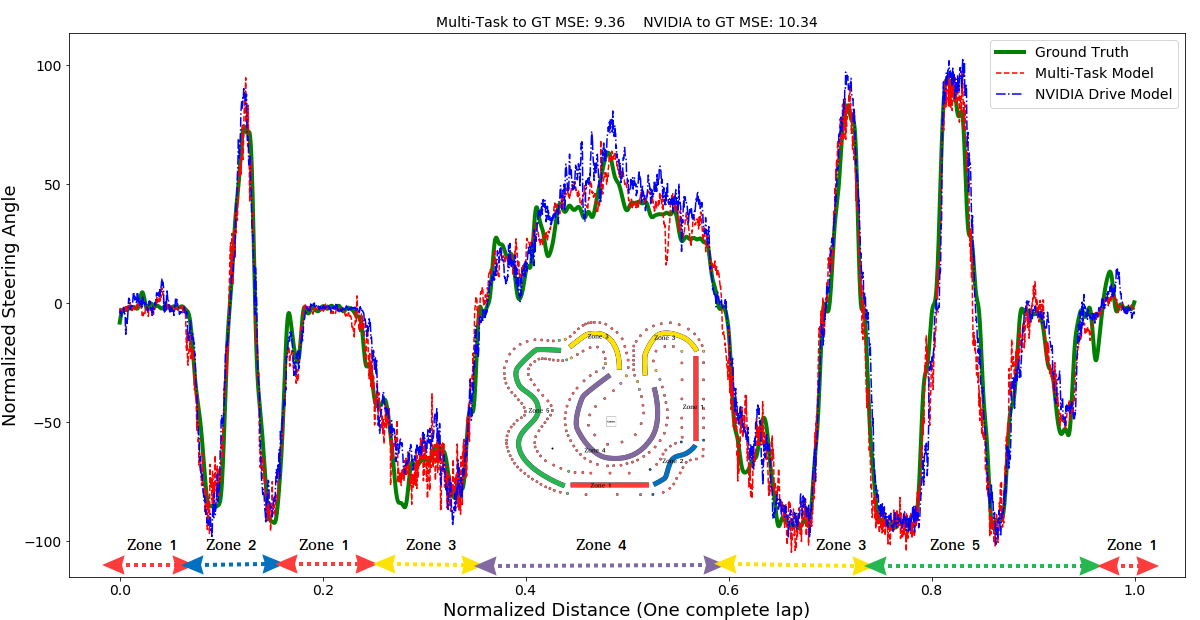}}
  \caption{Mult-task steering prediction based on test log files. The model presented in \cite{NVIDIA} is presented as well as a reference benchmark. Performance in terms of net MSE is roughly equivalent in both directions.}
  \label{fig:test_file_drive}
\end{figure}

When the MCN is coupled to the SRNs as illustrated in \figref{fig:two-level}, the predicted steering response for an entire lap can be produced. This is presented in \figref{fig:test_file_drive}. As a performance benchmark, the end-to-end driving model presented in \cite{NVIDIA} was trained with the same data set as used for the MCN as specified in \tableref{table:training_schedule}. (It is noted that the MCN used the zone labels in each data frame, while the model presented in \citep{NVIDIA} used the normalized steering values.)

As stated previously, in itself running inference on a test data set is not sufficient to ensure model quality, since the \textit{recovery} aspect of the model is not tested. In certain instances of an actual driving experiment, the vehicle will find itself removed from trajectories documented in either training or test sets, and as such, the models overall robustness is evaluated by monitoring if the vehicle can restore itself and maintain a consistent trajectory around the lap. This requires actual driving experiments with the model running inference on the scaled vehicle.

\newpage
\subsection{Empirical Driving Experiment}

\begin{figure}[htp]
  \centering
  \subfigure[Clockwise drive direction]{\label{cw_test}\includegraphics[scale=0.325]{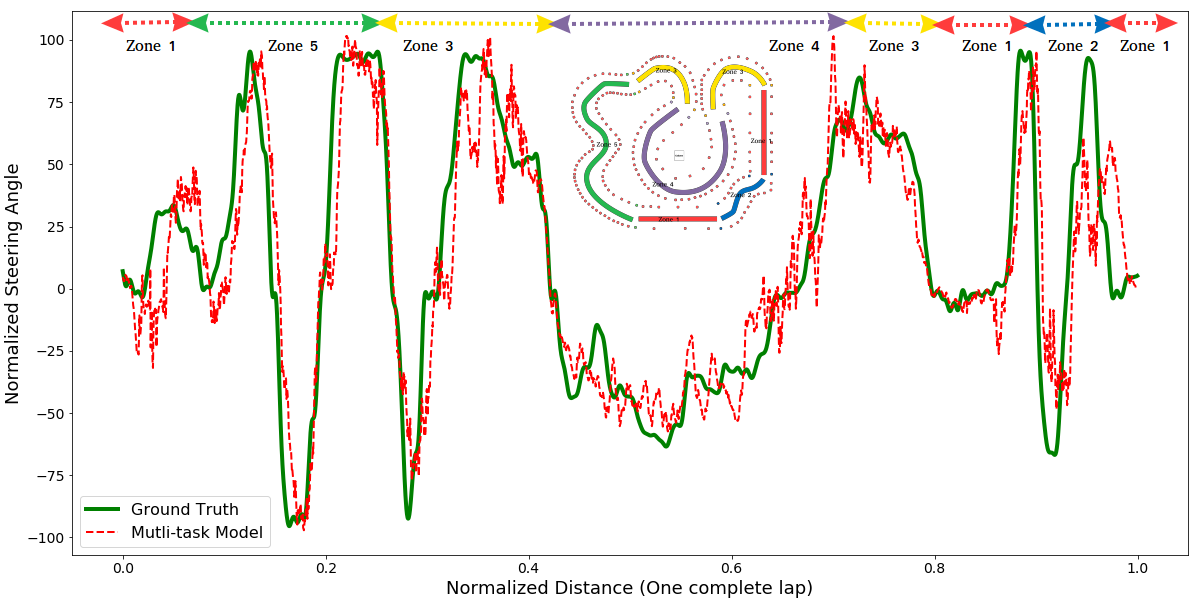}}
  \subfigure[Counter-clockwise drive direction]{\label{ccw_test}\includegraphics[scale=0.325]{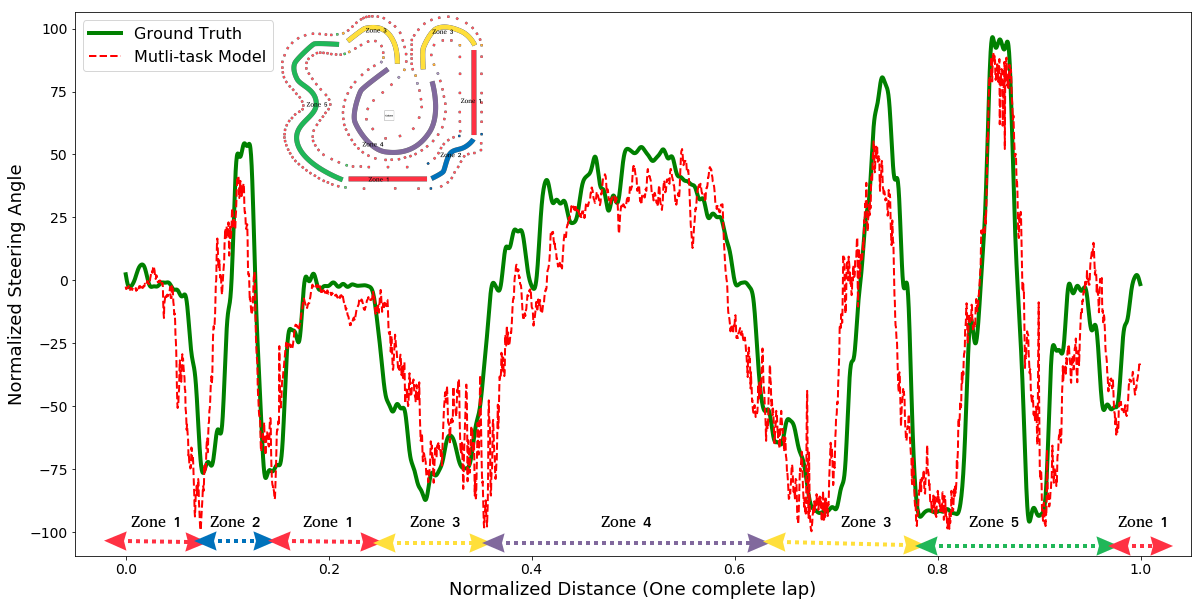}}
  \caption{Test drive around track comparing multi-task DNN architecture and a reference training data drive, (recorded from human driver input). Due to fluctuations in the frame rate grab logged by the vehicle while it is in autonomous mode, the normalized distance traveled while in autonomous mode was expanded by a fixed scalar value: 2\% in the clockwise direction and 3\% in the counter-clockwise direction. This was done to make a one-to-one comparison of the steering characteristics of both curves at the same point in the lap trajectory.}
  \label{fig:drive_experiment}
\end{figure}

The vehicle receives the trained weights of the MCN and SRNs, and then is put in autonomous mode to complete the lap in both directions. As mentioned previously, the standard for success is the completion of three laps in both directions without moving a cone. To quantify performance, the steering and video capture is logged while the vehicle is completing the task. Shown in \figref{fig:drive_experiment} is the steering output for the model in one of the three laps in each direction. A video recording of the vehicle completing a lap has been posted on \href{https://youtu.be/mBOEnStGNwo}{YouTube} \footnote{Video link: https://youtu.be/mBOEnStGNwo}. A ground truth steering profile is provided for reference, which is the steering logged from a human driver completing the lap.

As seen in the steering profiles, the vehicle completes the lap in both directions successfully. Although the log files document a fair amount of noise in the steering signal, this is not actually transferred to the steering mechanism since the inference frequency occurs at a much higher rate than the steering update cycle, thus the vehicle traverses the lap smoothly. 

By using the MCN to determine drive zones, it is noted that different throttle speeds where used for different driving zones in actual driving experiments. In watching the posted video, it can be seen that the vehicle speeds through straightaways but slows down for the chicane and tight turns. This component of controlling throttle based on driving task is a facet of the research that will be investigated further.

\section{Future Work}

The next steps in the current research is to implement the framework illustrated in \figref{fig:multiTaskUserIntent} and to move the research platform to the outside and experiment in realistic driving environments. Using the controlled setting of the five zone track allowed for the direct assessment of the MCN-SRN hierarchical model. With that component firmly established, the work of correlating user intent to the MCN framework can begin in earnest. In moving to real world external driving conditions, more direct applicability will result from the driving models developed from this platform to those of full-scale vehicles, (i.e. actual cars), currently under development as test platforms. The current work has created a framework for further experiments on a wide variety of road geometries and conditions that can be done using full-scale vehicles and in real-world conditions.

\section{Conclusion}

The multi-task model is a framework for tackling the complex task of E2E in real world, full-scale applications. Although a classical single network model could complete the navigation task of the five zone track using the same data set as used to train the MCN, it is important to note that Zone 1, 4 and 5 used much reduced training sets to create robust model performance. This is a pivotal contribution of the work given that it allows for tailored network design for distinct driving tasks, and couples model design to specified training to produce model robustness. In addition, the groundwork has been established for the user-intent framework that will be implemented based on the infrastructure developed by the current effort.

\section{Acknowledgments}

The authors would like to acknowledge Dr.~Dimitar Filev, Dr.~Gint Puskorius and Dr.~Dragos Maciuca for their assistance in setting up the research project. Their support at various stages was pivotal in conducting the work. The authors are also grateful to Dr.~Karl Zipser and Dr.~Sascha Hornauer for their input at the early stages of the project and introducing the authors to scaled vehicles as versatile instruments with which to conduct deep learning research.

\newpage
\bibliographystyle{plain}
\bibliography{Multi_Task_DNN_For_E2E_Driving}

\end{document}